\documentclass[letterpaper]{article} 
\usepackage{aaai24}  
\usepackage{times}  
\usepackage{helvet}  
\usepackage{courier}  
\usepackage[hyphens]{url}  
\usepackage{graphicx} 
\usepackage{natbib}  
\usepackage{caption} 
\usepackage{algorithm}
\usepackage{algorithmic}
\usepackage{amsmath}
\usepackage{amssymb}
\usepackage{braket}
\usepackage{wasysym}
\usepackage[frozencache=true,cachedir=minted-output]{minted}
\usepackage{newfloat}
\usepackage{listings}
\usepackage{multirow}
\usepackage{booktabs}
\usepackage{subcaption}
\DeclareCaptionStyle{ruled}{labelfont=normalfont,labelsep=colon,strut=off} 
\floatstyle{ruled}
\newfloat{listing}{tb}{lst}{}
\floatname{listing}{Listing}
\title{Narrowing the Gap between Supervised and Unsupervised \\Sentence Representation Learning with Large Language Model}
\author{
    Mingxin Li\textsuperscript{\rm 1}, Richong Zhang\textsuperscript{\rm 1,2}\thanks{Corresponding author}, Zhijie Nie\textsuperscript{\rm 1,3}, Yongyi Mao\textsuperscript{\rm 4}
}
\affiliations{
    \textsuperscript{\rm 1}SKLSDE, School of Computer Science and Engineering, Beihang University, Beijing, China\\
    \textsuperscript{\rm 2}Zhongguancun Laboratory, Beijing, China\\
    \textsuperscript{\rm 3}Shen Yuan Honors College, Beihang University, Beijing, China\\
    \textsuperscript{\rm 4}School of Electrical Engineering and Computer Science, University of Ottawa, Ottawa, Canada\\
    \{limx,zhangrc,niezj\}@act.buaa.edu.cn, ymao@uottawa.ca
}

\usepackage{bibentry}

\begin{document}

\maketitle

\begin{abstract}

Sentence Representation Learning (SRL) is a fundamental task in Natural Language Processing (NLP), with the Contrastive Learning of Sentence Embeddings (CSE) being the mainstream technique due to its superior performance. An intriguing phenomenon in CSE is the significant performance gap between supervised and unsupervised methods, 
with their only difference lying in the training data.
Previous works attribute this performance gap to differences in two representation properties (alignment and uniformity). However, since alignment and uniformity only measure the results, they fail to answer 
``\textit{What} aspects of the training data contribute to the performance gap?''
and ``\textit{How} can the performance gap be narrowed?''. 
In this paper, we conduct empirical experiments to answer these ``\textit{What}'' and ``\textit{How}'' questions. 
We first answer the ``\textit{What}'' question by thoroughly comparing the behavior of supervised and unsupervised CSE during their respective training processes. From the comparison, we identify the similarity pattern as a key factor to the performance gap, and introduce a metric, called \textit{Relative Fitting Difficulty} (RFD), to measure the complexity of the similarity pattern.
Then, based on the insights gained from the ``\textit{What}'' question, we tackle the ``\textit{How}'' question by increasing the pattern complexity of the training data. We achieve this by leveraging the In-Context Learning (ICL) capability of the Large Language Model (LLM) to generate data that simulates complex patterns. 
By utilizing the hierarchical patterns in the LLM-generated data, we effectively narrow the gap between supervised and unsupervised CSE. We release our codes and appendix at \underline{https://github.com/BDBC-KG-NLP/NGCSE}.

\end{abstract}

\begin{figure}[ht] 
    \centering
    \includegraphics[width=1.0\linewidth]{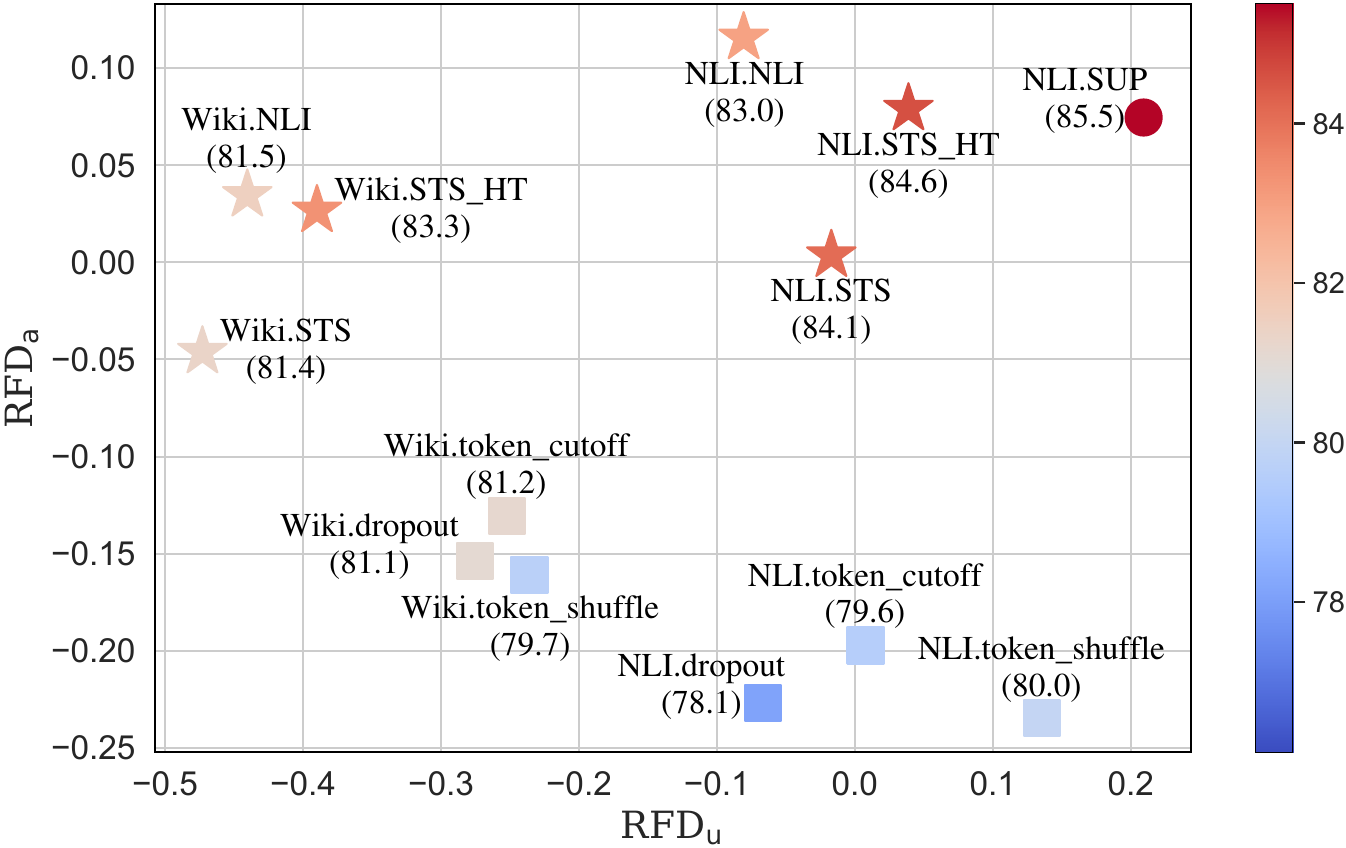} 
    \caption{RFD$_\mathrm{u}$-RFD$_\mathrm{a}$ plot of models based on BERT$_\mathrm{base}$. The colors of the points and the numbers in brackets represent Spearman's correlation in the evaluation data, i.e., the validation split of STS Benchmark dataset. 
    RFD is a metric we propose to measure the complexity of the similarity pattern in the training data.
    This metric helps us answer the ``\textit{What}'' question and further leads us to address the ``\textit{How}'' question. ``$\blacksquare$'' is trained with unsupervised data, ``$\CIRCLE$'' is trained with supervised data, and ``$\bigstar$'' is trained with data generated from the LLM.}
    \label{fig:rfd}
\end{figure}

\section{Introduction}

\begin{figure}[ht] 
    \centering
    \includegraphics[width=1.0\linewidth]{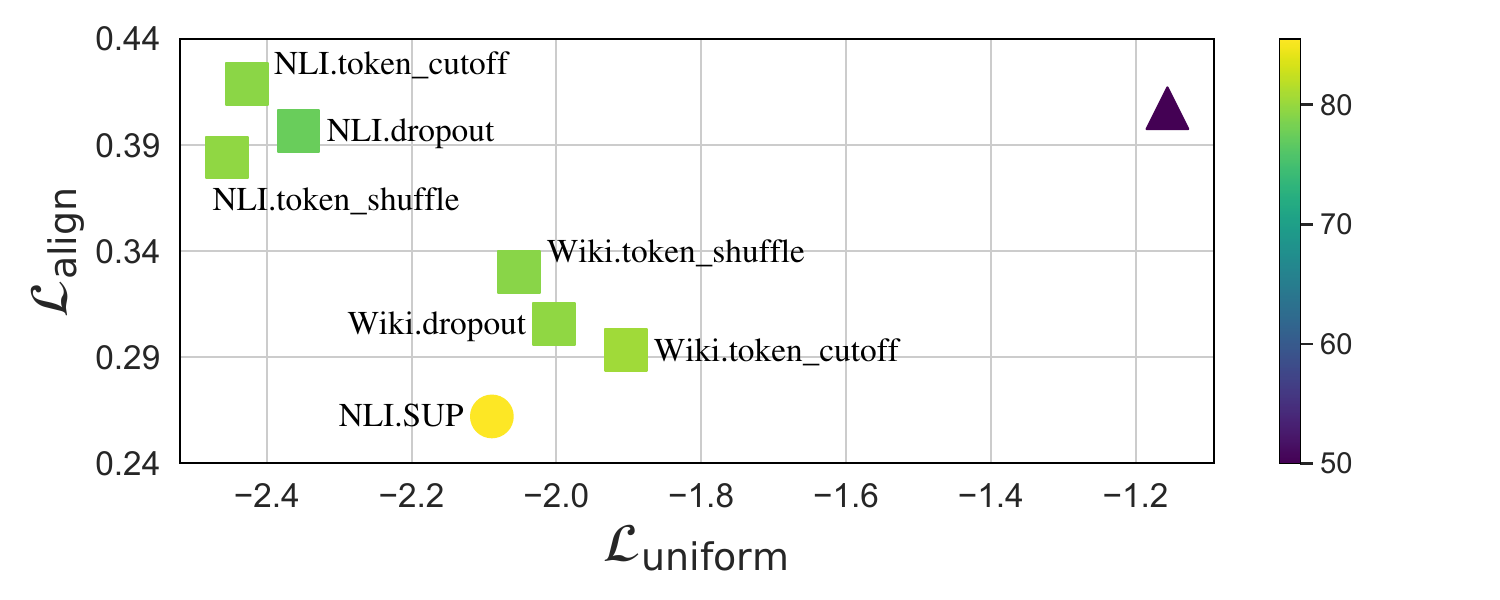} 
    \caption{Uniformity-alignment plot of models based on BERT$_\mathrm{base}$. The colors of the points and the numbers in brackets represent Spearman's correlation in the validation split of STS Benchmark dataset. ``$\blacktriangle$'' is the pre-trained model, ``$\blacksquare$'' is trained with unsupervised data, and ``$\CIRCLE$'' is trained with supervised data.
    }
    \label{fig:align_uniform}
\end{figure}

Sentence Representation Learning (SRL) is a crucial task in Natural Language Processing (NLP), which learns representations (or embeddings) for sentences in the feature space. It is a fundamental task that underpins many NLP applications, including Semantic Textual Similarity (STS)~\cite{wang_understanding_2020}, Information Retrieval (IR)~\cite{cer_universal_2018}, and text classification~\cite{pang_sentimental_2004}. 
Contrastive learning of Sentence Embeddings (CSE) has been recently introduced into SRL~\cite{yan_consert_2021,gao_simcse_2021} and has drawn much attention as it significantly improves the performance of sentence embeddings. CSE can be trained in both supervised and unsupervised settings, where the primary difference is the training data. However, with only this difference, supervised CSE can outperform unsupervised CSE on STS tasks by a large margin. \citet{gao_simcse_2021} explain this performance gap 
by referring to two properties (alignment and uniformity) from \cite{wang_understanding_2020}. Specifically, compared to the representations trained by the unsupervised method, they find that the representations trained by supervised data exhibit better alignment, uniformity or both (as shown in Figure~\ref{fig:align_uniform}), 
thereby resulting in better performance on STS tasks. 
This explanation still rests on the final results, and cannot explain the mechanism that led to these results. In this paper, we focus on the training data and its relationship with the performance gap.
Specifically, we pose two questions: ``\textit{What} aspects of the training data contribute to the performance gap?'' and ``\textit{How} can the performance gap be narrowed?'', and answer them with empirical experiments in this study.

To answer the ``\textit{What}'' question, we record the variety of alignment and uniformity in the training processes of both supervised and unsupervised CSE, where we identify the similarity pattern, i.e., how a dataset defines similar and dissimilar sentence pairs, as a key factor to the performance gap. The more complex the similarity pattern of a training dataset, the higher the performance that training with such a dataset can yield. We also find that the complexity of the similarity pattern (pattern complexity for short) can be measured by the relative magnitude of alignment and uniformity between the training data and the evaluation data. 
More specifically, the similarity pattern of the supervised training data is more difficult to fit than that of the evaluation data, resulting in higher alignment and uniformity values in the training data than those in the evaluation data. In contrast, the similarity pattern of the unsupervised training data is simpler to fit, resulting in lower alignment and uniformity values. 
Therefore, we define a metric called \textit{Relative Fitting Difficulty} (RFD) to measure the pattern complexity and provide the answer to the ``\textit{What}'' question: the increase of the pattern complexity leads to the performance gap between supervised and unsupervised CSE.

Based on the insight gained from answering the ``\textit{What}'' question, we answer the ``\textit{How}'' question by introducing complex similarity patterns into the unsupervised training data. This is achieved by leveraging the In-Context Learning (ICL) capability of the Large Language Model (LLM)~\cite{brown_language_2020} to simulate the similarity patterns in STS~\cite{agirre_semeval-2012_2012} and NLI~\cite{gao_simcse_2021} datasets. Furthermore, we notice the hierarchical nature of the STS dataset, where the semantic similarity between two sentences is measured with a score ranging from 0 to 5, rather than simply classified as similar or dissimilar. This finding motivates us to simulate the hierarchical pattern of the STS dataset. And to utilize the hierarchical pattern, we propose a loss called Hierarchical Triplet (HT) loss to ensure that such a pattern can be learned during training, which helps us further narrow the performance gap. 

Briefly, our main contributions are as follows:
\begin{itemize}
    \item We propose a new metric, i.e., \textit{Relative Fitting Difficulty} (RFD), to measure the complexity of the similarity pattern
    and demonstrate that the higher RFDs on both alignment and uniformity correlate with better performance on STS tasks;
    \item We narrow the performance gap on STS tasks between supervised and unsupervised CSE by introducing the training data with complex similarity patterns, which is obtained by the ICL capability of LLMs, and introduce a novel loss function called Hierarchical Triplet (HT) loss to utilize the hierarchical patterns effectively;
    \item We conduct extensive further experiments to validate our findings on RFDs and to verify the effectiveness of our proposed methods in narrowing the performance gap.
\end{itemize}

\section{Background}
\label{sec:background}
In this section, we first detail the differences in training and performance between supervised and unsupervised CSE to help readers understand the background better, and then give some notations for the convenience of the later narrative.

\subsection{Performance Gap on the STS tasks}


Contrastive learning of Sentence Embeddings (CSE)~\cite{yan_consert_2021, gao_simcse_2021} is a prevalent technique in SRL for its superior performance. CSE refines the sentence embeddings in two ways: 1) pulling the anchor sentence $s_i$ and its semantically similar sentence (or \textbf{positive sentence} $s_i^\mathrm{p}$) closer; 2) pushing the anchor sentence $s_i$ and its semantically dissimilar sentence (or \textbf{negative sentence} $s_i^\mathrm{n}$) apart.
The commonly used contrastive learning loss, InfoNCE \cite{oord_representation_2018}, can be expressed as
\begin{align}\label{eq:CL}
    \mathcal{L}_\mathrm{C} =
    -\log\frac{e^{f(s_i)^\top f(s_i^\mathrm{p})/\tau}}{e^{f(s_i)\top f(s_i^\mathrm{p})/\tau}+\sum_{j=1}^Ne^{f(s_i)^\top f(s_{i,j}^\mathrm{n})/\tau}},
\end{align}
where $\tau$ is a hyper-parameter and $\{s_{i,j}^\mathrm{n}\}_{j=1}^N$ is a set of negative sentences corresponding to $s_i$.

There are two settings of CSE, namely Supervised CSE (S-CSE) and Unsupervised CSE (U-CSE). Many improvements have been proposed based on both S-CSE and U-CSE~\cite{jiang_promptbert_2022,wu_infocse_2022}, but we focus on the typical paradigm in this study. 
For U-CSE, the augmented view of $s_i$ is treated as its positive sentence $s_i^\mathrm{p}$, and a randomly sampled sentence is treated as its negative sentence $s_i^\mathrm{n}$. The common data augmentation methods used to generate this augmented view include dropout, token shuffle, and token cutoff~\cite{yan_consert_2021}. U-CSE is typically trained with a Wikipedia dataset~\cite{gao_simcse_2021}, consisting of one million sentences extracted from Wikipedia. For S-CSE, it relies on human-annotated $s_i^\mathrm{p}$ and $s_i^\mathrm{n}$, with the supervision signal from the Natural Language Inference (NLI) task~\cite{conneau_supervised_2017}. 
Specifically, given a premise, the entailment hypothesis is treated as $s_i^\mathrm{p}$ and the contradiction hypothesis is treated as $s_i^\mathrm{n}$. \citet{gao_simcse_2021} collect a widely used NLI dataset~\cite{gao_simcse_2021} for CSE.

The performance of CSE is usually evaluated with the SentEval toolkit~\cite{conneau_senteval_2018}, which includes the STS tasks and the transfer tasks. The STS task quantifies the semantic similarity between two sentences with a score ranging from 0 to 5 and takes Spearman's correlation as the metric for performance. There are seven STS datasets are included for evaluation: STS 2012-2016~\cite{agirre_semeval-2012_2012, agirre_sem_2013, agirre_semeval-2014_2014, agirre_semeval-2015_2015, agirre_semeval-2016_2016}, STS Benchmark~\cite{cer_semeval-2017_2017}, SICK Relatedness~\cite{marelli_sick_2014}. The transfer tasks evaluate the transfer capability of sentence embeddings by performing logistic regression. There are also seven datasets included for the evaluation of transfer task: MR~\cite{pang_seeing_2005}, CR~\cite{hu_mining_2004}, SUBJ~\cite{pang_sentimental_2004}, MPQA~\cite{wiebe_annotating_2005}, SST-2~\cite{socher_recursive_2013}, TREC~\cite{voorhees_building_2000}, and MRPC~\cite{dolan_automatically_2005}. 

Both S-CSE and U-CSE exhibit strong performance in the transfer tasks, but there exists a significant performance gap in the STS task, even when the sole difference between them lies in the training data. \citet{gao_simcse_2021} borrow two properties, alignment and uniformity, from the empirical work of \citet{wang_understanding_2020} to  better understand it, and the two properties can be expressed as
\begin{align}
    & \mathcal{L}_{\mathrm{align}}\triangleq\underset{(s,s^\mathrm{p})\sim p_\mathrm{pos}}{\mathbb{E}}\|f(s)-f(s^\mathrm{p})\|^\alpha_2\\
    & \mathcal{L}_{\mathrm{uniform}}\triangleq\log\underset{(s_i,s_j)\stackrel{{\text{{i.i.d.}}}}{\sim}p_\mathrm{data}}{\mathbb{E}}e^{-t\|f(s_i)-f(s_j)\|^2_2}
\end{align}
where $\alpha$ and $t$ are two hyper-parameters, $p_\mathrm{pos}$ is the distribution of positive sentence pairs, and $p_\mathrm{data}$ is the distribution of sentences. These properties measure the quality of the sentence embeddings.
\citet{gao_simcse_2021} has shown that the performance gap results from the better alignment and uniformity of S-CSE in comparison to U-CSE (as shown in Figure~\ref{fig:align_uniform}).
This explains the result of the performance gap, but does not shed light on the question of ``\textit{What} aspects of the training data contribute to the performance gap?''. In this study, we seek to answer this ``\textit{What}'' question. Furthermore, based on the insights from the ``\textit{What}'' question, we will explore ``\textit{How} can the performance gap be narrowed?''.

\subsection{Notation}

We conduct experiments under various training data settings to study how the training data affects the performance of CSE. To maintain consistency, we organize all settings using the same naming format: ``\textbf{[Data-Domain].[Similarity-Pattern]}'', 
where [Data-Domain] represents how the anchor sentences are collected, and [Similarity-Pattern] represents how the positive and negative sentences are defined. Two data domains are included: (1) Wiki, consisting of sentences from Wikipedia~\cite{gao_simcse_2021}; (2) NLI, consisting of the premises from the NLI dataset~\cite{gao_simcse_2021}. And similarity patterns are divided into three types, including supervision signals (denoted as SUP), data augmentations, and our proposed pattern simulation techniques, which will be explained in later sections.

\section{\textit{What} aspects of the training data contribute to the performance gap?}
\label{sec:what}




\begin{figure}[ht] 
    \centering
    \includegraphics[width=1.0\linewidth]{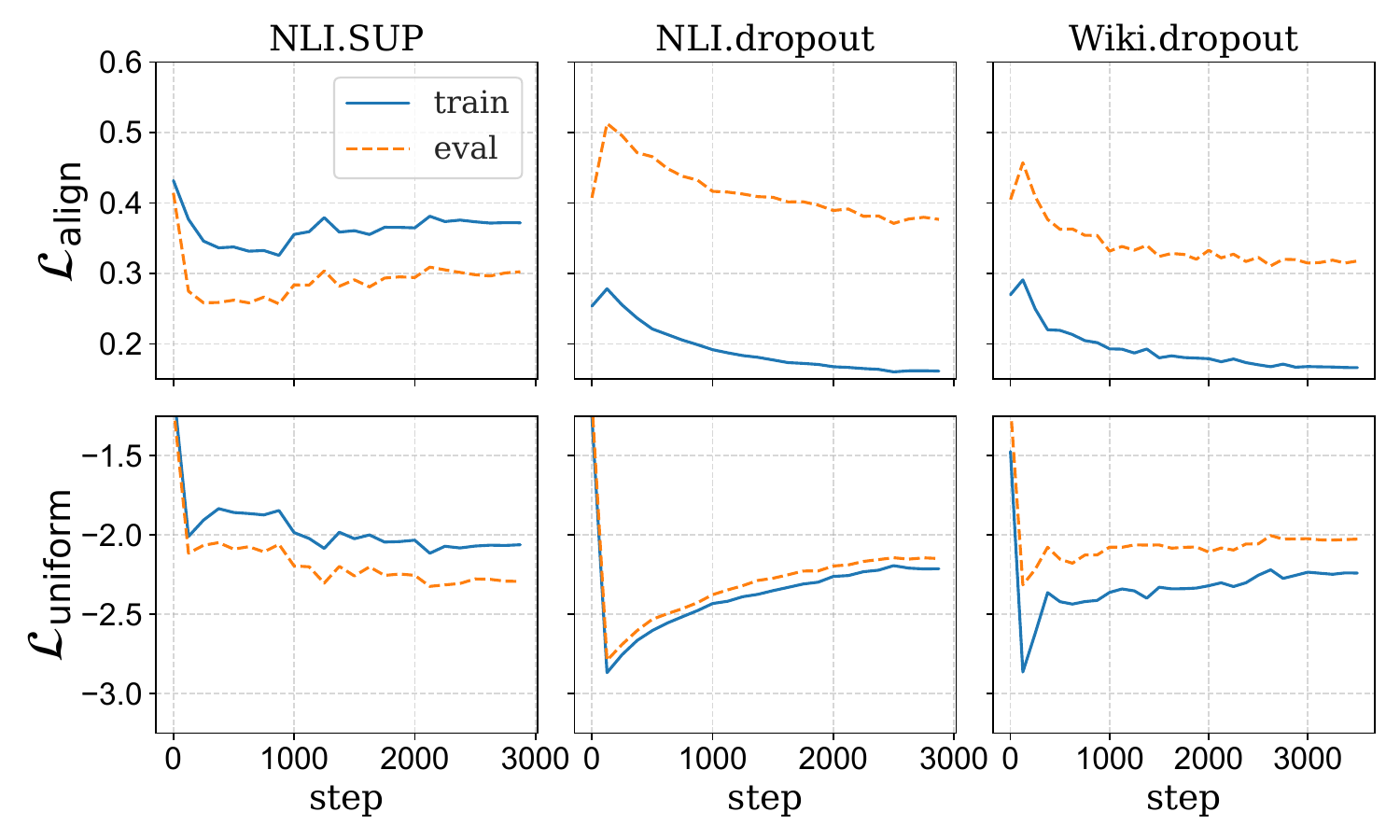} 
    \caption{Alignment and uniformity in both the held-out training data and the evaluation data during the training process. We only plot the results of ``NLI.SUP'', ``NLI.dropout'', and ``Wiki.dropout'' here for comparison. The results of ``token\_shuffle'' and ``token\_cutoff'' are similar to those of ``dropout'', so we plot them in the \textbf{appendix}.}
    \label{fig:align_uniform_process}
\end{figure}

\subsection{Observation in the Training Process}
\label{sec:observation}

In this section, we study how the training data affects the performance of CSE. To ensure the results are only correlated with training data, we select pre-trained BERT$_\mathrm{base}$ model~\cite{devlin_bert_2019} as the backbone and ensure identical settings across all training, with details illustrated in the \textbf{appendix}. Then, we record changes in alignment and uniformity in both the training and evaluation data during the training process and observe how these changes vary between S-CSE and U-CSE. Note that the widely-used training data for S-CSE (``NLI.SUP'') and U-CSE (``Wiki.dropout'', ``Wiki.token\_cutoff'', ``Wiki.token\_shuffle'') differ in both their data domains and similarity patterns. To study the impact of these two factors separately, we introduce three additional training data settings to U-CSE: ``NLI.dropout'', ``NLI.token\_cutoff'', and ``NLI.token\_shuffle''. Part of the results are shown in Figure~\ref{fig:align_uniform_process}.

We first fix the similarity pattern to investigate the impact of the data domain, e.g., comparing ``NLI.dropout'' with ``Wiki.dropout''. The results in Figures~\ref{fig:align_uniform} and \ref{fig:align_uniform_process} show that the performance of the STS task and the trends in alignment and uniformity are similar across different data domains, indicating that the data domain does not influence the performance significantly. 
Next, we fix the data domain to investigate the impact of the similarity pattern, e.g., comparing ``NLI.SUP'' with ``NIP.dropout''. From the comparison, we can observe the performance gap in Figure~\ref{fig:align_uniform}, indicating that the similarity pattern is a key factor in such phenomenon. Furthermore, Figure~\ref{fig:align_uniform_process} shows that the changes in alignment and uniformity during the training process in S-CSE are quite different from those in U-CSE. Specifically, S-CSE exhibits higher alignment and uniformity values in the held-out training data than those in the evaluation data. In contrast, U-CSE exhibits lower alignment and uniformity values in the held-out training data than those in the evaluation data. We argue that this difference in the training process, in fact, reflects the difference in the complexity of the similarity pattern (pattern complexity for short).


\subsection{Relative Fitting Difficulty}

The similarity pattern in S-CSE, which is defined by supervision signals, is far more complex than the similarity pattern in U-CSE, which is defined by data augmentations. Moreover, the similarity pattern in S-CSE training data is more difficult to fit than that of the evaluation data, while the similarity pattern in U-CSE training data is simpler to fit than that of the evaluation data. This difference in the pattern complexity results in the difference in the training process between S-CSE and U-CSE. Therefore, we introduce a metric called \textit{Relative Fitting Difficulty} (RFD) to act as an indicator of the \textbf{pattern complexity}. RFD is defined as the difference in fitting difficulty between the held-out training data and the evaluation data, i.e., the relative magnitude of alignment and uniformity between the held-out training data and the evaluation data during the training process. 

Let the alignment and uniformity of a sentence encoder $f$ at time step $t$ in the held-out training data be denoted by $a_\mathrm{h}(f, t)$ and $u_\mathrm{h}(f, t)$, while those in the evaluation data be denoted by $a_\mathrm{e}(f, t)$ and $u_\mathrm{e}(f, t)$. We can then define the RFD for alignment over a set of time steps $T = \{t_i\}_{i=1}^M$ as
\begin{align}
\mathrm{RFD_a}(f, T) &= \frac{1}{M}\sum_{i=1}^M a_\mathrm{h}(f, t_i) - a_\mathrm{e}(f, t_i),
\end{align}
and the RFD for uniformity as
\begin{align}
\mathrm{RFD_u}(f, T) &= \frac{1}{M}\sum_{i=1}^M u_\mathrm{h}(f, t_i) - u_\mathrm{e}(f, t_i).
\end{align}
We calculate the RFD for the six U-CSE settings and one S-CSE setting mentioned in the last subsection, and present the results in Figure~\ref{fig:rfd} using ``$\blacksquare$'' and ``$\CIRCLE$''. 
By comparing the results within each data domain, we can observe two facts: (1) when one setting has both lower $\mathrm{RFD_a}$ and $\mathrm{RFD_u}$ values than another setting, it will have lower performance in the STS task accordingly; (2) When either $\mathrm{RFD_a}$ or $\mathrm{RFD_u}$ value increases, the performance in the STS task tends to improve.
These observations show a tendency that higher RFD values correspond to better performance. In other words, compared to U-CSE, S-CSE has higher fitting difficulty in alignment and uniformity, i.e., higher pattern complexity, which leads to better performance in the STS task. In fact, we conjecture it may be the answer to the what question. 

However, this answer is drawn from only seven points in two data domains. If we need to get this answer more conclusive, we need to experiment under more settings and to get more RFD coordinates and their corresponding STS task performance. Therefore, in the next section, we will introduce some artificial settings and explore the correlation between their RFDs and STS performance to further corroborate the conclusions of this section.







\section{\textit{How} can the performance gap be narrowed?}

In this section, we answer the ``\textit{How}'' based on the insights gained from the ``\textit{What}''. At the same time, we will provide additional validation for our answer to the ``\textit{What}'' question.


\begin{figure*}
  \centering
  \includegraphics[width=1.\textwidth]{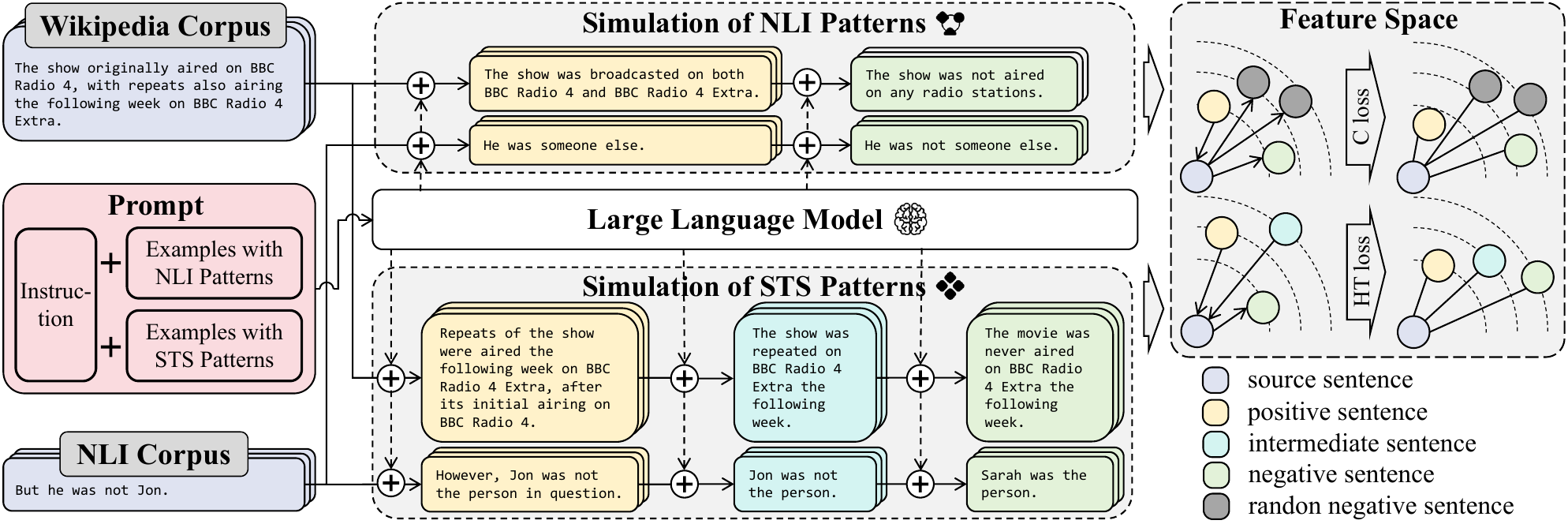}
  \caption{The procedure of pattern simulation and pattern utilization. We simulate the STS and NLI patterns separately with the ICL capability of the LLM. Then, we adopt a combination of contrastive loss and HT loss proposed by us to utilize the pattern. The prompt consists of three examples randomly sampled from the pattern source and its detail is shown in the \textbf{appendix}.}
  \label{fig:procedure}
\end{figure*}

\subsection{Pattern Simulation with LLM}
\label{sec:simulation}
Our answer to the ``\textit{What}'' question reveals the correlation between the pattern complexity in training data and the STS task performance. Therefore, to narrow the performance gap, we propose to increase the RFD of U-CSE by introducing complex similarity patterns (patterns for short) into U-CSE.
To realize this, we leverage the In-Context Learning (ICL)~\cite{brown_language_2020} capability of LLM to simulate the patterns in NLI and STS datasets.
We adopt the \texttt{gpt-3.5-turbo-0613} as the LLM through the official API~\footnote{https://openai.com/api} from OpenAI with default parameters.

Figure~\ref{fig:procedure} illustrates the overall procedure of pattern simulation and dataset generation. The datasets are generated from two types of sources: a corpus source and a pattern source. For the corpus source, we consider the two data domains used in the previous experiments: (1) Wiki, which consists of sentences from Wikipedia, and (2) NLI, which consists of the premises from the NLI dataset. For the pattern source, we also consider two classes of patterns: (1) STS patterns, which adopt the training split of the STS12 dataset~\cite{agirre_semeval-2012_2012} as the source of patterns, and (2) NLI patterns, which adopt the same NLI dataset as previously as the source patterns. 

When generating datasets, we randomly sample 20,000 sentences from the corpus source and use the LLM to simulate STS and NLI patterns separately. We refer to these sentences as source sentences. For every source sentence $s_i$,  we subsequently generate a positive sentence $s_i^\mathrm{p}$ and a negative sentence $s_i^\mathrm{n}$ using the following process:
\begin{enumerate}
    \item \textbf{Sampling Pattern Example}:
    To simulate STS patterns, We randomly sample three sentence pairs with STS scores above 4 as examples for the pattern of positive sentence pairs (referred to as positive examples), and three sentence pairs with STS scores below 1 as examples for the pattern of negative sentence pairs (referred to as negative examples). Similarly, to simulate NLI patterns, we randomly sample three premises and their entailment hypotheses as positive examples, and three premises and their contradiction hypotheses as negative examples;
    \item \textbf{Generating Positive Sentence $s_i^\mathrm{p}$}:
    To generate $s_i^\mathrm{p}$ that simulates STS patterns, we prompt the LLM with positive examples to generate a sentence that is semantically similar to $s_i$. And to generate $s_i^\mathrm{p}$ that simulate NLI patterns, we prompt the LLM with positive examples to generate a sentence that is an entailment hypothesis to $s_i$;
    \item \textbf{Generating Negative Sentence $s_i^\mathrm{n}$}:
    To generate $s_i^\mathrm{n}$ that simulates STS patterns, we prompt the LLM with negative examples to generate a sentence with a distinct meaning compared to $s_i^\mathrm{p}$. And to generate $s_i^\mathrm{n}$ that simulate NLI patterns, we prompt the LLM with negative examples to generate a sentence that contradicts $s_i^\mathrm{p}$.
\end{enumerate}
The data generated in these processes is called ``LLM-generated data'' for simplicity. We combine the LLM-generated data with the remaining sentences in the source corpus to form hybrid datasets, resulting in a total of four hybrid datasets. We name the training data settings of these four datasets by: ``Wiki.STS'', ``Wiki.NLI'', ``NLI.STS'', and ``NLI.NLI''.  Note that (1) For ``Wiki.NLI'' and ``NLI.NLI'', the supervision signals in the NLI dataset are used; (2) For ``NLI.STS'', we only utilize the premises, which do not contain any supervision signals in the NLI dataset. 

Then we employ the four training data settings to perform CSE
under the same setting as in the \textbf{Observation} section. This allows us to examine whether the introduced complex patterns can help narrow the performance gap between S-CSE and U-CSE. Additionally, we calculate both $\mathrm{RFD}_{a}$ and $\mathrm{RFD}_{u}$ for all four settings to further validate our answer to the ``\textit{What}'' question. The results are plotted in Figure~\ref{fig:rfd} using ``$\bigstar$''. From the results, it is evident that all the results trained by the hybrid datasets outperform the U-CSE, indicating success in narrowing the performance gap. Moreover, all of these new settings exhibit larger RFD values ($\mathrm{RFD_a}$ or $\mathrm{RFD_u}$ or both) compared to U-CSE,
which indicates that we indeed introduce complex patterns to the training data that lead to an increase in performance.
Also, these observations can be viewed as evidence to support our answer to the``\textit{What}'' question.

\begin{table*}[ht]\scriptsize
\centering
\begin{tabular}{lccccccccc}
\toprule
\textbf{Method} & \textbf{Group} & \textbf{STS12} & \textbf{STS13} & \textbf{STS14} & \textbf{STS15} & \textbf{STS16} & \textbf{STS-B} & \textbf{SICK-R} & \textbf{Avg.} \\
\midrule
InferSent\textsuperscript{$\dagger$} & \multirow{5}{*}{
    \begin{tabular}{@{}c@{}}
      \textit{Supervised} \\
      \textit{methods}
    \end{tabular}
} & 52.86 & 66.75 & 62.15 & 72.77 & 66.87 & 68.03 & 65.65 & 65.01 \\
SBERT\textsuperscript{$\dagger$} & & 70.97 & 76.53 & 73.19 & 79.09 & 74.30 & 77.03 & 72.91 & 74.89 \\
ConSERT\textsuperscript{*} & & 74.07 & 83.93 & 77.05 & 83.66 & 78.76 & 81.36 & 76.77 & 79.37 \\
SimCSE\textsuperscript{*} & & 75.30 & 84.67 & 80.19 & 85.40 & 80.82 & 84.25 & 80.39 &  81.57 \\
PromptBERT\textsuperscript{$\dagger$} & & \textbf{75.48} & \textbf{85.59} & \textbf{80.57} & \textbf{85.99} & \textbf{81.08} & \textbf{84.56} & \textbf{80.52} & \textbf{81.97} \\
\midrule
BERT-whitening\textsuperscript{$\ddagger$} & \multirow{5}{*}{
\begin{tabular}{@{}c@{}}
      \textit{Unsupervised} \\
      \textit{methods}
    \end{tabular}
} & 57.83 & 66.90 & 60.90 & 75.08 & 71.31 & 68.24 & 63.73 & 66.28 \\
ConSERT\textsuperscript{*} & & 64.64 & 78.49 & 69.07 & 79.72 & 75.95 & 73.97 & 67.31 & 72.74 \\
SimCSE\textsuperscript{*} & & 68.40 &  82.41 & 74.38 & 80.91 & 78.56 & 76.85 & 72.23 &  76.25 \\
PromptBERT\textsuperscript{$\dagger$} & & 71.56 & 84.58 & 76.98 & 84.47 & 80.60 & 81.60 & 69.87 & 78.54 \\
InfoCSE\textsuperscript{$\ddagger$} & & 70.53 & 84.59 & 76.40 & \underline{85.10} & \underline{81.95} & 82.00 & 71.37 & 78.85 \\
\midrule
Dino\textsuperscript{\S} & \multirow{2}{*}{
    \begin{tabular}{@{}c@{}}
      \textit{LM-based} \\
      \textit{methods}
    \end{tabular}
} & 72.61 & 81.92 & 75.09 & 80.42 & 76.26 & 77.10 & 70.43 & 76.26 \\
CLAIF\textsuperscript{\S} & & 70.62 & 81.51 & 76.29 & 85.05 & 81.36 & \underline{84.34} & \underline{78.22} & 79.63 \\
\midrule
Wiki.STS\_HT & \multirow{2}{*}{
\begin{tabular}{@{}c@{}}
      \textit{Our} \\
      \textit{methods}
    \end{tabular}
} & 72.46$_\mathrm{\pm 0.15}$  & \underline{84.88}$_\mathrm{\pm 0.40}$ & \underline{77.80}$_\mathrm{\pm 0.63}$ & 83.85$_\mathrm{\pm 0.66}$ & 81.11$_\mathrm{\pm 0.44}$ & 81.90$_\mathrm{\pm 0.18}$ & 76.56$_\mathrm{\pm 0.26}$ & 79.79$_\mathrm{\pm 0.23}$ \\
NLI.STS\_HT & & \underline{72.94}$_\mathrm{\pm 0.19}$ & 84.32$_\mathrm{\pm 0.27}$ & 77.71$_\mathrm{\pm 0.29}$ & 84.20$_\mathrm{\pm 0.40}$ & 80.85$_\mathrm{\pm 0.27}$ & 82.21$_\mathrm{\pm 0.19}$ & 78.04$_\mathrm{\pm 0.35}$ & \underline{80.02}$_\mathrm{\pm 0.22}$ \\
\bottomrule
\end{tabular}
\caption{Spearman's correlation on STS Tasks. All models adopt \textbf{BERT}$_\mathrm{base}$ as the backbone. $\dagger$: results from \cite{jiang_promptbert_2022}, $\ddagger$: results from \cite{wu_infocse_2022}, \S: results from \cite{cheng_improving_2023}, *: results from their original paper. We bold the highest results among all models and underline the highest results among the models that are not supervised.}
\label{tab:sts}
\end{table*}

\begin{table*}[ht]\scriptsize
\centering
\begin{tabular}{lccccccccc}
\toprule
\textbf{Method} & \textbf{Group} & \textbf{STS12} & \textbf{STS13} & \textbf{STS14} & \textbf{STS15} & \textbf{STS16} & \textbf{STS-B} & \textbf{SICK-R} & \textbf{Avg.} \\
\midrule
SRoBERTa\textsuperscript{$\dagger$} & \multirow{3}{*}{
    \begin{tabular}{@{}c@{}}
      \textit{Supervised} \\
      \textit{methods}
    \end{tabular}
} & 71.54 & 72.49 & 70.80 & 78.74 & 73.69 & 77.77 & 74.46 & 74.21 \\
SimCSE\textsuperscript{*} & & 76.53 & 85.21 & 80.95 & 86.03 & 82.57 & 85.83 & 80.50 & 82.52 \\
PromptRoBERTa\textsuperscript{$\dagger$} & & \textbf{76.75} & \textbf{85.93} & \textbf{82.28} & \textbf{86.69} & 82.80 & \textbf{86.14} & \textbf{80.04} & \textbf{82.95} \\
\midrule
RoBERTa-whitening\textsuperscript{$\dagger$} & \multirow{3}{*}{
\begin{tabular}{@{}c@{}}
      \textit{Unsupervised} \\
      \textit{methods}
    \end{tabular}
} & 57.83 & 63.24 & 57.23 & 71.36 & 68.99 & 61.36 & 62.91 & 61.73 \\
SimCSE\textsuperscript{*} & & 70.16 &  81.77 & 73.24 & 81.36 & 80.65 & 80.22 & 68.56 &  76.57 \\
PromptRoBERTa\textsuperscript{$\dagger$} & & 73.94 & 84.74 & 77.28 & 84.99 & 81.74 & 81.88 & 69.50 & 79.15 \\
\midrule
Dino\textsuperscript{\S} & \multirow{2}{*}{
    \begin{tabular}{@{}c@{}}
      \textit{LM-based} \\
      \textit{methods}
    \end{tabular}
} & 71.24 & 81.55 & 75.67 & 81.42 & 78.77 & 80.10 & 71.31 & 77.15 \\
CLAIF\textsuperscript{\S} & & 68.33 & 82.26 &  77.00 & 85.18 & \underline{\textbf{83.43}} & \underline{85.05} & 78.02 & 79.90 \\
\midrule
Wiki.STS\_HT & \multirow{2}{*}{
\begin{tabular}{@{}c@{}}
      \textit{Our} \\
      \textit{methods}
    \end{tabular}
} & \underline{75.68}$_\mathrm{\pm 0.41}$ & 84.97$_\mathrm{\pm 0.63}$ & 78.08$_\mathrm{\pm 0.69}$ & 84.82$_\mathrm{\pm 0.23}$ & 83.41$_\mathrm{\pm 0.37}$ & 83.79$_\mathrm{\pm 0.25}$ & 77.66$_\mathrm{\pm 0.20}$ & 81.20$_\mathrm{\pm 0.21}$ \\
NLI.STS\_HT & & 74.54$_\mathrm{\pm 0.51}$  & \underline{85.10}$_\mathrm{\pm 0.42}$ & \underline{79.10}$_\mathrm{\pm 0.15}$ & \underline{85.48}$_\mathrm{\pm 0.19}$ & 82.93$_\mathrm{\pm 0.28}$ & 83.87$_\mathrm{\pm 0.17}$ & \underline{78.31}$_\mathrm{\pm 0.27}$ & \underline{81.33}$_\mathrm{\pm 0.09}$ \\
\bottomrule
\end{tabular}
\caption{Spearman's correlation on STS Tasks. All models adopt \textbf{RoBERTa}$_\mathrm{base}$ as the backbone. $\dagger$: results from \cite{jiang_promptbert_2022}, \S: results from \cite{cheng_improving_2023}, *: results from their original paper. We bold the highest results among all models and underline the highest results among the models that are not supervised.}
\label{tab:sts_roberta}
\end{table*}

\subsection{Pattern Utilization with Hierarchical Triplet Loss}

In the previous subsection, we have managed to narrow the performance gap to some extent. However, there is still something not being fully utilized, which is the hierarchical nature of the STS patterns. Instead of defining the positive sentence pair and negative sentence pair, the STS task adopts a score ranging from 0 to 5 to reflect the semantic similarity between two sentences, which makes the similarity pattern in STS dataset hierarchical. To maintain such hierarchical nature of STS patterns, we revise our process of pattern simulation (as shown in Figure~\ref{fig:procedure}). Specifically, we prompt the LLM to generate an intermediate sentence $s_i^\mathrm{m}$ which contains the less details compared to the positive sentence $s_i^\mathrm{p}$,
and we randomly sample three sentences from the source of STS patterns with STS scores between 1 and 4 as examples in the prompt. Then, we propose a method to utilize all three sentences $s_i^\mathrm{p}$, $s_i^\mathrm{m}$ and $s_i^\mathrm{n}$ by adopting a sequence of triplet losses. This approach ensures that the hierarchical pattern can be learned by the sentence encoder. We refer to this loss as the Hierarchical Triplet (HT) loss, and we provide its formal definition below.

For a source sentence $s_i$ and the three sentences generated based on it $s_i^\mathrm{p}$, $s_i^\mathrm{m}$ and $s_i^\mathrm{n}$, the HT loss is defined as
\begin{align}
    \mathcal{L}_\mathrm{HT} = \frac{1}{2}(&\max(f(s_i)^\top f(s_i^\mathrm{m}) - f(s_i)^\top f( s_i^\mathrm{p}) + m_1, 0) + \notag\\
    &\max(f(s_i)^\top f(s_i^\mathrm{n}) - f(s_i)^\top f(s_i^\mathrm{m}) + m_2, 0)), 
\end{align}
where $m_1, m_2$ are two hyper-parameters that control the margin of the triplet loss, and $f$ is the sentence encoder that maps sentences into a hypersphere.
The HT loss is combined with the contrastive loss~\ref{eq:CL} to form the final loss:
\begin{align}
    \mathcal{L} = \mathcal{L}_\mathrm{C} + \beta \mathcal{L}_\mathrm{HT}, 
\end{align}
where $\beta$ is a hyper-parameter controls the weight of $\mathcal{L}_\mathrm{HT}$. Note that $\mathcal{L}_\mathrm{HT}$ is calculated only on the LLM-generated data, which covers 20,000 instances in the hybrid dataset. 

We perform CSE with this final loss on ``Wiki.STS'' and ``NLI.STS'' under the same setting as in the \textbf{Observation} section, These training settings are denoted as ``Wiki.STS\_HT'' and ``NLI.STS\_HT'' respectively. For all settings in this section, we set $m_1=5e-3$, $m_2=1e-2$ and $\beta=1$. Similarly, we calculate the RFD of these settings and plot the results in Figure~\ref{fig:rfd} using $\bigstar$ points. It can be observed that training with $\mathcal{L}_{\mathrm{HL}}$ increases both $\mathrm{RFD_a}$ and $\mathrm{RFD_u}$, and then improves the performance.
The rise of RFD values can be explained as follows: the common pattern only determines which sentence pair is similar and dissimilar, while the hierarchical pattern determines which sentence pair is more similar and dissimilar than another. In other words, the hierarchical pattern extends the ideas of the common pattern, thereby increasing the pattern complexity and raising RFD values.

Through the above subsections, we significantly narrow the performance gap between S-CSE and U-CSE. We now provide our answer to the ``\textit{How}'' question: By utilizing the ICL capability of LLM, we can simulate the patterns in the NLI and STS datasets, thereby introducing complex patterns to the unsupervised training dataset. This process narrows the performance gap to some extent. Subsequently, we thoroughly exploit the hierarchical patterns in the STS dataset with the HT loss, further narrowing the performance gap.

\subsection{Final Performance}
\label{sec:final_performance}

In this section, we compare our methods with various well-known and state-of-the-art baselines:

\noindent\textbf{Unsupervised baselines} include a post-processing method, BERT-whitening~\cite{su_whitening_2021}, as well as contrastive learning methods like ConSERT~\cite{yan_consert_2021}, SimCSE~\cite{gao_simcse_2021},PromptBERT~\cite{jiang_promptbert_2022}, and InfoCSE~\cite{wu_infocse_2022}.

\noindent\textbf{Supervised baselines} include some traditional supervised methods such as InferSent~\cite{conneau_supervised_2017}, SBERT~\cite{reimers_sentence-bert_2019}, and some of the contrastive learning methods mentioned above, which can also be utilized in a supervised setting.

\noindent\textbf{LM-based baselines} includes Dino~\cite{schick_generating_2021}, which generates training data with the Pre-trained Language Model (PLM), 
and CLAIF~\cite{cheng_improving_2023}, which generates training data with the LLM.

In the previous sections, all experiments were conducted under the same settings for a fair comparison.  While in this section, we run the ``Wiki.STS\_HT'' and ``NLI.STS\_HT'' training settings under a group of hyper-parameters and decide the best combination of hyper-parameters with the evaluation data, i.e., the validation split STS Benchmark dataset. The details are provided in the \textbf{appendix}.
We get sentence embeddings following \citet{jiang_promptbert_2022} to achieve better and more stable performance.
We evaluate our method following the standard evaluation protocol mentioned in the \textbf{Background} section and compare our method with the baselines on the STS tasks in Table~\ref{tab:sts} (with BERT$_\mathrm{base}$ as backbone) and \ref{tab:sts_roberta} (with RoBERTa$_\mathrm{base}$~\cite{liu_roberta_2019} as backbone). 
By comparing our method with both supervised and unsupervised baselines, we observe that although our method is still inferior to state-of-the-art supervised methods, it outperforms all unsupervised baselines by a large margin. This indicates that we successfully narrow the performance gap between supervised and unsupervised CSE. When compared to data-generation-based methods, our method outperforms them by generating only 20,000 instances, which is significantly fewer than them. 

\section{Further Study}

In this section, we investigate how each component of our method impacts the performance, and how the embeddings transfer to the downstream tasks. We conduct the experiments on ``Wiki.STS\_HT'' with BERT$_\mathrm{base}$ as the backbone.

\begin{figure}[ht]
    \centering
    \begin{subfigure}[b]{0.47\textwidth}
        \centering
        \includegraphics[width=\textwidth]{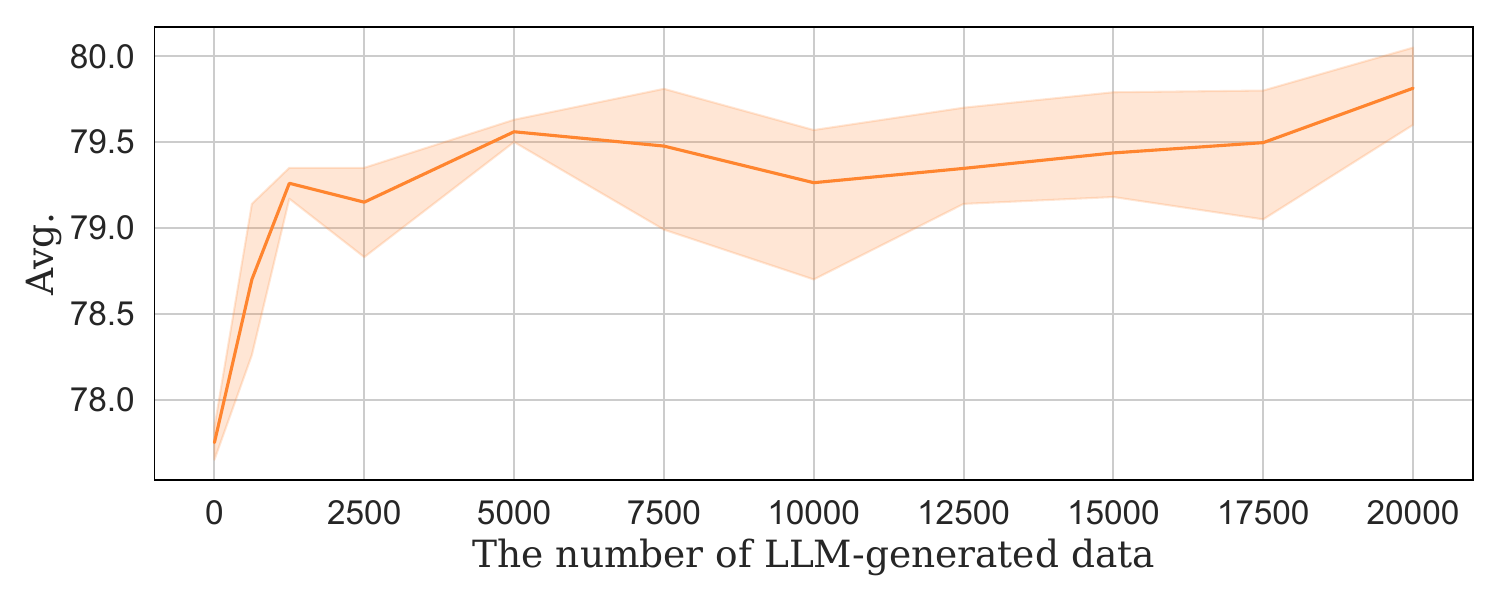}
        \caption{The average Spearman's correlation on STS tasks w.r.t the number of LLM-generated data in the hybrid training dataset.}
        \label{fig:number}
    \end{subfigure}

    \begin{subfigure}[b]{0.47\textwidth}
        \centering
        \includegraphics[width=\textwidth]{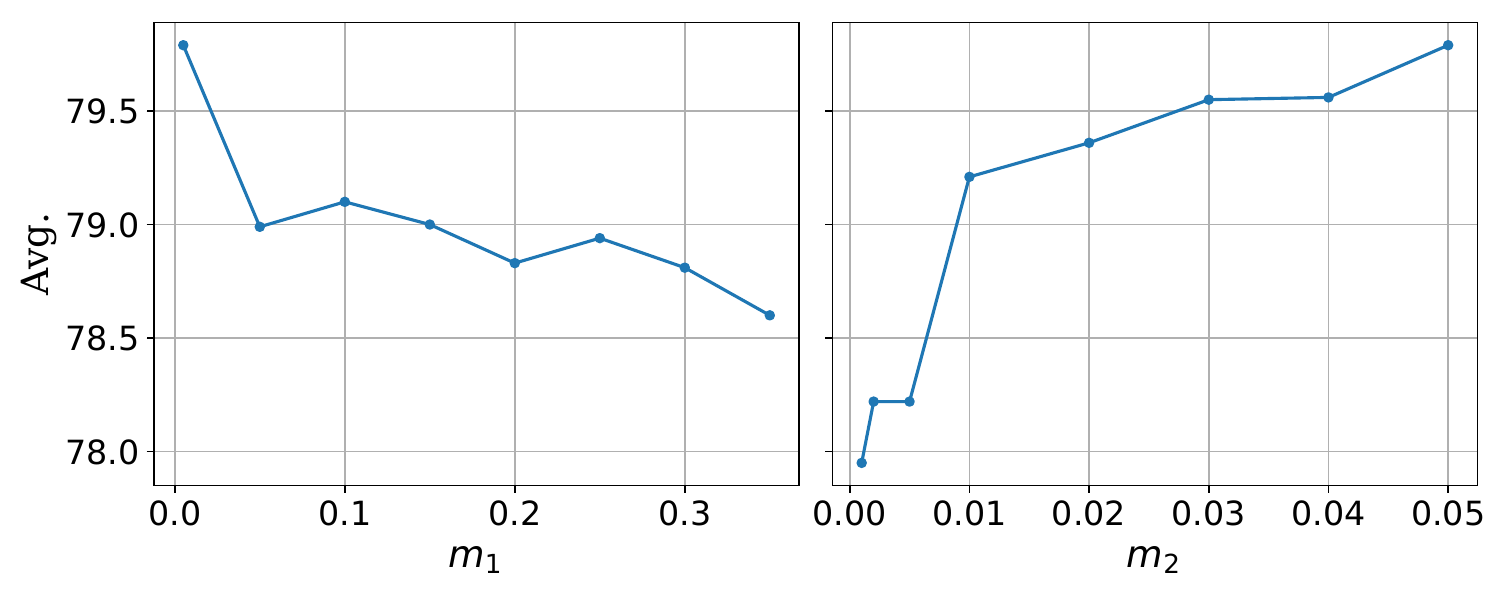}
        \caption{The average Spearman's correlation on STS tasks w.r.t the margins ($m_1$ and $m_2$) in HT loss.}
        \label{fig:m1_m2}
    \end{subfigure}
    \caption{The study of hyper-parameters.}
    \label{fig:hp_study}
\end{figure}

\subsection{The Number of LLM-generated data}


In this section, we want to investigate the impact of the number, which is 20,000 previously, of LLM-generated data on performance. To this end, We run our method under different numbers of LLM-generated data and plot the results in Figure~\ref{fig:number}. From this figure, we can observe a scaling effect that as the number of LLM-generated data increases, the performance of our method tends to improve. 

\subsection{The Margins of HT Loss}


There are two margins in the HT loss, $m_1$ and $m_2$. In pattern simulation, both $s_i^\mathrm{p}$ and $s_i^\mathrm{m}$ can be regarded as positive samples to the source sentence. As such, we set a small $m_1$ to ensure that the distance between $f(s_i^\mathrm{p})$ and $f(s_i^\mathrm{m})$ is not large when learning their hierarchical pattern. Conversely, $s_i^\mathrm{n}$ can be treated as a negative sample to the source sentence, so we set a large $m_2$ to ensure that the distance between $f(s_i^\mathrm{m})$ and $f(s_i^\mathrm{n})$ is sufficiently large when learning their hierarchical pattern. In this section, we invert these settings to investigate the impact of a large $m_1$ and small $m_2$ on performance. The results are plotted in Figure~\ref{fig:m1_m2}, and they conform to our expectation that large $m_1$ and small $m_2$ would adversely affect the performance.

\subsection{HT Loss v.s. Contrastive Loss}

The HT loss is proposed to ensure that the sentence encoder can learn the hierarchical STS pattern. However, training with the HT loss includes more positive samples (i.e., $s_i^\mathrm{m}$) than training without the HT loss. To investigate whether the improvement in performance brought about by the HT loss is solely due to the presence of more positive samples, we run our method with only the contrastive loss in two settings: \textit{single positive}, where only $s_i^\mathrm{p}$ is treated as positive samples, and \textit{multiple positive}, while both $s_i^\mathrm{p}$ and $s_i^\mathrm{m}$ are treated as positive samples. The results are shown in Table~\ref{tab:comparison}. ``CL.\textit{multiple positive}'' outperforms ``CL.\textit{single positive}'', indicating that more positive samples indeed improve performance. However, it still underperforms when compared to training with the HT loss, suggesting that the HT loss indeed brings more effective signals to the training process.

\begin{table}[htbp]
  \centering
    \begin{tabular}{lc}
    \toprule
          & \textbf{Avg.} \\
    \midrule
    Wiki.STS   & \textbf{79.79}$_\mathrm{\pm 0.23}$  \\
    CL.\textit{single\_psitive}    & 78.72$_\mathrm{\pm 0.24}$  \\
    CL.\textit{multiple\_positive}    & 79.14$_\mathrm{\pm 0.27}$  \\
    \bottomrule
    \end{tabular}%
  \caption{The average Spearman's correlation on STS tasks when HT loss is replaced with Contrastive Loss (CL).
  }
  \label{tab:comparison}%
\end{table}%

\subsection{Transfer Tasks}

Our study focuses on improving sentence embeddings in the STS task performance, so it remains a mystery how well the learned sentence embeddings can be applied to the downstream tasks. In this subsection, we evaluate the performance of transfer tasks for our methods following the standard evaluation protocol mentioned in the \textbf{Background} section, and compare our methods with the baselines in Table~\ref{tab:transfer_tasks}. In addition to the baselines mentioned in the \textbf{Final Perfomance} section, we also adopt two pooling strategies of BERT as baselines in this subsection. 
The statistics in the table show that our methods outperform the methods that are not supervised on most of the tasks, and exhibits competitive performance compared to the supervised methods. These results indicate that, by improving the STS task performance, our methods can learn sentence embeddings that are suitable for the downstream tasks.

\begin{table*}[htbp]\scriptsize
  \centering
    \begin{tabular}{lcccccccc}
    \toprule
    \textbf{Model} & \textbf{MR} & \textbf{CR} & \textbf{SUBJ} & \textbf{MPQA} & \textbf{SST} & \textbf{TREC} & \textbf{MRPC} & \textbf{Avg.} \\
    \midrule
    \midrule
    \multicolumn{9}{c}{\textit{Supervised methods}} \\
    \midrule
    InferSent\textsuperscript{$\dagger$} & 81.57  & 86.54  & 92.50  & \textbf{90.38}  & 84.18  & 88.20  & 75.77  & 85.59  \\
    SBERT\textsuperscript{$\dagger$} & \textbf{83.64}  & \textbf{89.43}  & 94.39  & 89.86  & \textbf{88.96}  & 89.60  & 76.00  & \textbf{87.41}  \\
    SimCSE\textsuperscript{*} & 82.69  & 89.25  & 94.81  & 89.59  & 87.31  & 88.40  & 73.51  & 86.51  \\
    PromptBERT\textsuperscript{$\dagger$} & 83.14  & 89.38  & 94.49  & 89.93  & 87.37  & 87.40  & 76.58  & 86.90  \\
    \midrule
    \midrule
    \multicolumn{9}{c}{\textit{Unsupervised methods}} \\
    \midrule
    Avg. BERT embedding\textsuperscript{$\dagger$} & 78.66 &  86.25 &  94.37  & 88.66 & 84.40 & \underline{\textbf{92.80}} & 69.54 & 84.94 \\
    BERT-[CLS] embedding\textsuperscript{$\dagger$} &  78.68 & 84.85 & 94.21 &  88.23 & 84.13 & 91.40 & 71.13 & 84.66 \\
    SimCSE\textsuperscript{*} & 81.18  & 86.46  & 94.45  & 88.88  & 85.50  & 89.80  & 74.43  & 85.81  \\
    PromptBERT\textsuperscript{$\dagger$} & 80.74  & 85.49  & 93.65  & 89.32  & 84.95  & 88.20  & 76.06  & 85.49  \\
    InfoCSE\textsuperscript{*} & 81.76  & 86.57  & \underline{\textbf{94.90}}  & 88.86  & 87.15  & 90.60  & 76.58  & 86.63  \\
    \midrule
    \midrule
    \multicolumn{9}{c}{\textit{LM-based methods}} \\
    \midrule
    Dino\textsuperscript{\S}  & 79.96  & 85.27  & 93.67  & 88.87  & 84.29  & 88.60  & 69.62  & 84.33  \\
    CLAIF\textsuperscript{\S} & 81.64  & 87.98  & 94.24  & 89.34  & 86.16  & 89.80  & \underline{\textbf{77.16}}  & 86.62  \\
    \midrule
    \midrule
    \multicolumn{9}{c}{\textit{Our methods}} \\
    \midrule
    Wiki.STS\_HT & 82.12$_\mathrm{\pm 0.63 
}$  & 87.96$_\mathrm{\pm 0.27 }$  & 94.82$_\mathrm{\pm 0.18 
}$  & 90.10$_\mathrm{\pm 0.13 }$  & 86.84$_\mathrm{\pm 1.38 
}$  & 89.20$_\mathrm{\pm 1.74 }$  & 75.88$_\mathrm{\pm 1.81 
}$  & 86.70$_\mathrm{\pm 0.76 }$  \\
    NLI.STS\_HT & \underline{82.36}$_\mathrm{\pm 0.61 }$  & \underline{88.19}$_\mathrm{\pm 0.11 }$  & 94.62$_\mathrm{\pm 0.15 }$  & \underline{90.15}$_\mathrm{\pm 0.19 }$  & \underline{87.75}$_\mathrm{\pm 0.45 }$  & 90.60$_\mathrm{\pm 0.60 }$  & 76.93$_\mathrm{\pm 0.53 }$  & \underline{87.23}$_\mathrm{\pm 0.24 }$  \\
    \bottomrule
    \end{tabular}%
    \caption{
    Results on the transfer tasks. All models adopt BERT$_\mathrm{base}$ as the backbone. We use the names of hybrid datasets to denote our models. $\dagger$: results from \cite{jiang_promptbert_2022}, \S: results from \cite{cheng_improving_2023}, *: results from their original paper. We bold the highest results among all models and underline the highest results among the models that are not supervised.}
  \label{tab:transfer_tasks}%
\end{table*}%

\section{Related Work}

Sentence Representation Learning (SRL)~\cite{conneau_supervised_2017,reimers_sentence-bert_2019,li_sentence_2020} is a fundamental task in NLP, aiming to learn representations for sentences that maintain semantic information.  The supervised~\cite{conneau_supervised_2017,reimers_sentence-bert_2019} and unsupervised~\cite{li_sentence_2020,su_whitening_2021} settings of SRL used to diverge a lot, where supervised SRL~\cite{conneau_supervised_2017,reimers_sentence-bert_2019} focused on how to utilize NLI datasets and unsupervised SRL~\cite{li_sentence_2020,su_whitening_2021} focused on how to mitigate the anisotropy problem~\cite{li_sentence_2020}. 
With the introduction of contrastive learning into SRL~\cite{yan_consert_2021,gao_simcse_2021}, many recent works~\cite{gao_simcse_2021,jiang_promptbert_2022} can be applied to both supervised and unsupervised SRL, building a bridge between these two settings. Although these works boost the performance of SRL under the contrastive learning paradigm, they do not explore the underlying processes leading to the performance gap between supervised CSE (S-CSE) and unsupervised CSE (U-CSE), which motivates our study. Our study also provides a method to improve U-CSE, which can be related to the studies~\cite{wu_infocse_2022} that specifically focus on U-CSE.


Our study utilizes the LLM in SRL, and a recent study by \cite{cheng_improving_2023} employs this approach as well. However, our study differs from theirs in both the method of data generation and the intention of using the LLM. They
generate data by predicting masks, while we do so by pattern simulation. They use the LLM to enhance the performance of CSE, while we use the LLM to narrow the performance gap between supervised and unsupervised CSE, and concurrently validate our findings about the fitting difficulty. There is also an early work~\cite{schick_generating_2021} that generates datasets with Pre-trained Language Models (PLM) in a way similar to ours. 
Though our methods seem similar, their intention is to mitigate the need for human-generated data, which is different from our intention.

To the best of our knowledge, we are the first to study what aspects of the training data contribute to the performance gap between supervised and unsupervised CSE.

\section{Conclusion}

In this study, we investigate the training process of S-CSE and U-CSE, where we find that the similarity pattern of training data is a key factor to the STS task performance. Then, we define a new metric called Relative Fitting Difficulty (RFD) to quantify the complexity of the similarity pattern in the training data,
and prove that higher RFD values correlate with improved performance.
Building on this insight, we successfully narrow the performance gap between S-CSE and U-CSE by introducing STS and NLI patterns to the unsupervised data.
Moreover, we introduce a Hierarchical Triplet (HT) loss to
utilize the hierarchical STS patterns, further narrowing the gap.
The fact that we train better sentence embeddings with hierarchical STS patterns than with NLI patterns indicates that a more advanced model may be trained by replacing the long-used NLI dataset with a carefully-crafted hierarchical STS dataset. Such a dataset, previously difficult to create due to the lack of sentences with hierarchical semantic similarities, is now attainable thanks to the powerful LLM.



\section{Acknowledgements}

This work was supported by the STI 2030-Major Projects under Grant 2022ZD0120200, in part by the National Natural Science Foundation of China (No. U23B2056), in part by the Fundamental Research Funds for the Central Universities, and in part by the State Key Laboratory of Software Development Environment.

\bibliography{aaai24}

\newpage
\clearpage

\section{Appendix}

\subsection{Training Details}

For the experiments before the \textbf{Final Performance} section, we conduct them under the same setting for a fair comparison. Specifically, we initialize a BERT~\cite{devlin_bert_2019} model from the pre-trained \texttt{bert-base-uncased} checkpoint and use the last hidden state of the \texttt{[CLS]} token as the sentence embedding. We train the model
under the hyper-parameters listed in Table~\ref{tab:hp_observation}. We run trainings on the Wiki data domain for one epoch, and on the NLI data domain for three epochs. We record the Spearman's correlation on the evaluation data, i.e., the validation split of STS Benchmark dataset, as well as the alignment and uniformity of both held-out training data and evaluation data, every 125 training steps. We randomly shuffle the original training data and hold out the top 10\% of the total data as the held-out training data. In Figure~\ref{fig:rfd}, the final score is the average of top 5 Spearman's correlation.

\begin{table}[htbp]
  \centering
    \begin{tabular}{cccc}
    \toprule
    Optimizer  & Learning rate
     & Batch size & Temp \\
    \midrule
    AdamW & 3e-5  & 256   & 5e-2 \\
    \bottomrule
    \end{tabular}%
  \caption{Hyper-parameters for experiments before the \textbf{Final Performance} section. AdamW is proposed by \cite{loshchilov_decoupled_2019} and Temp is the temperature $\tau$ for contrastive loss.}
  \label{tab:hp_observation}%
\end{table}%

For the experiments in the \textbf{Final Performance} sections, we adopt the pooling strategy from ~\cite{jiang_promptbert_2022} for the sentence embedding to achieve better and more stable performance. Specifically, the sentence $s$ is placed in the following prompts:
\begin{itemize}
    \item \textit{This sentence: ``s'' means [MASK]};
    \item \textit{This sentence of ``s'' means [MASK]}.
\end{itemize}
Then, the last hidden state of \texttt{[MASK]} is used as the sentence embedding. To determine the optimal combination of hyper-parameters, we retain the same backbone, optimizer, batch size and temp as the previous experiments, and carry out a grid search on the learning rate $\in \{1\mathrm{e}-5, 3\mathrm{e}-5, 5\mathrm{e}-5\}$, $m_1 \in \{1\mathrm{e}-3, 5\mathrm{e}-3, 1\mathrm{e}-2\}$, $m_2 \in \{1\mathrm{e}-2, 5\mathrm{e}-2, 1\mathrm{e}-1\}$ and $\beta \in \{0.1, 0.5, 1.0\}$ over the evaluation data, with the selected hyper-parameters listed in Table~\ref{tab:hp_final_performance}.

\begin{table}[htbp]
  \centering
    \begin{tabular}{ccccc}
    \toprule
          & Learning rate & $m_1$    & $m_2$    & $\beta$ \\
    \midrule
    Wiki.STS\_HT & 1e-5  & 5e-3  & 5e-2  & 1 \\
    NLI.STS\_HT & 1e-5  & 5e-3  & 1e-1  & 1 \\
    \bottomrule
    \end{tabular}%
  \caption{Hyper-parameters for experiments in the \textbf{Final Performance} section.}
  \label{tab:hp_final_performance}%
\end{table}%

Our code is implemented in \texttt{python 3.9.13}, with \texttt{pytorch 1.12.1}~\cite{paszke_pytorch_2019} and \texttt{transformers 4.18.0}~\cite{wolf_huggingfaces_2019}. The experiments are conducted on a single 32G NVIDIA V100 GPU and repeated three times with different random seeds to obtain the final results.

\subsection{Training Processes of ``token\_shuffle'' and ``token\_cutoff''}

\begin{figure*}[ht] 
    \centering
    \includegraphics[width=1.0\linewidth]{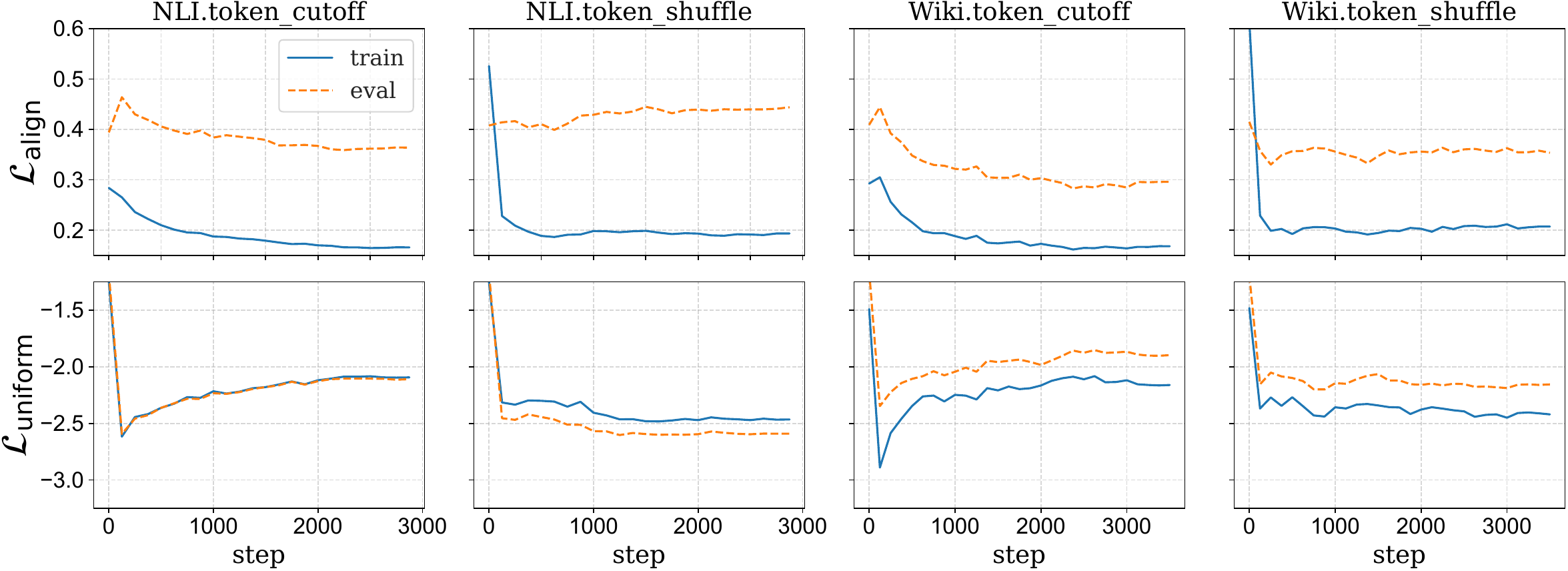} 
    \caption{Alignment and uniformity in both the held-out training data and the evaluation data during the training processes of ``NLI.token\_shuffle'', ``NLI.token\_cutoff'', ``Wiki.token\_shuffle'' and ``Wiki.token\_cutoff''.}
    \label{fig:align_uniform_process1}
\end{figure*}

We record the variety of alignment and uniformity during the training process in the \textbf{Observation} subsection. In this subsection, we present the results of ``NLI.token\_shuffle'', ``NLI.token\_cutoff'', ``Wiki.token\_shuffle'' and ``Wiki.token\_cutoff'' as shown in Figure~\ref{fig:align_uniform_process1}. Both exhibit better alignment and uniformity in the held-out training data compared to those in the evaluation data, which is similar to the training process of ``NLI.dropout'' and ``Wiki.dropout'' respectively.

\begin{figure*}
  \centering
  \includegraphics[width=1.\textwidth]{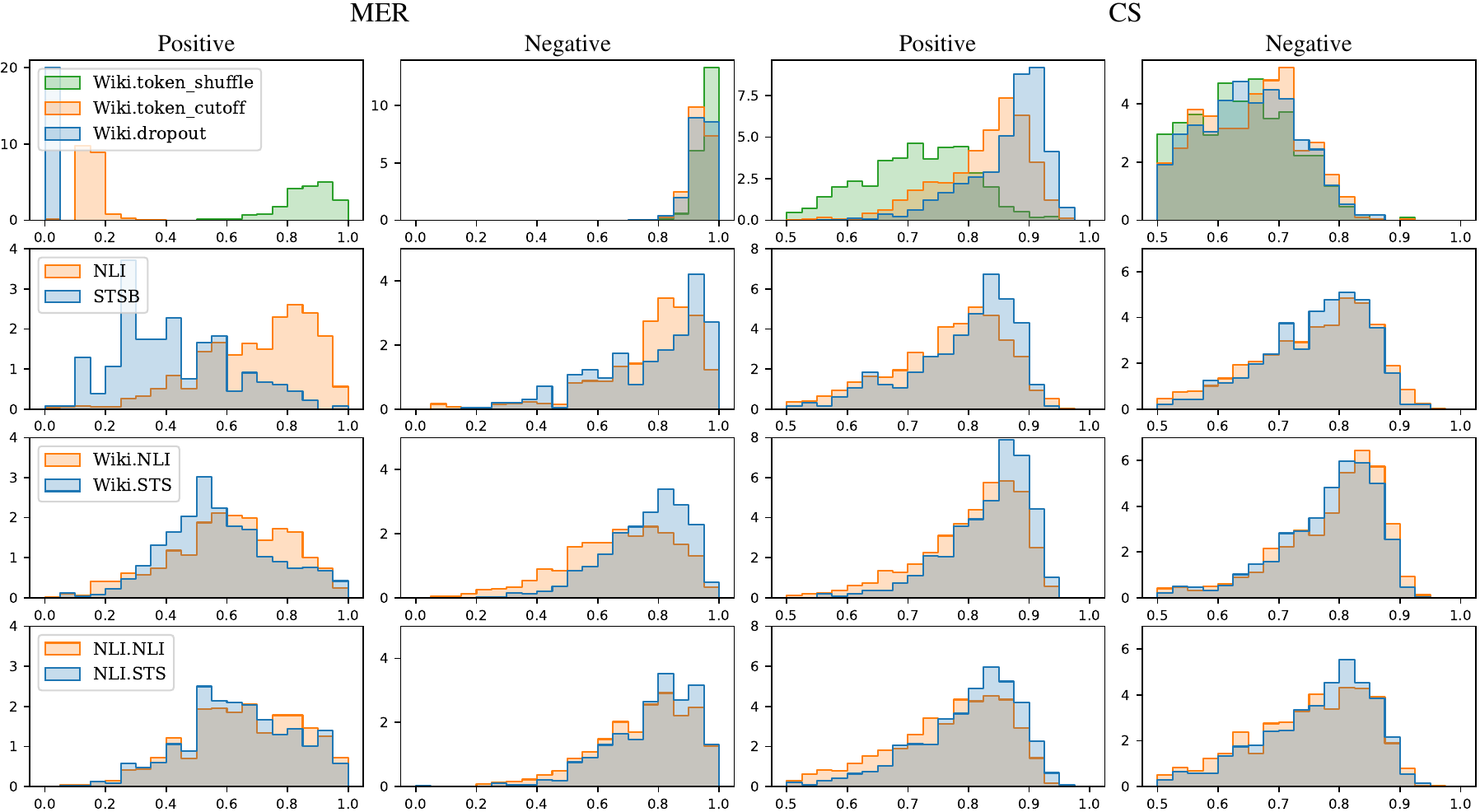}
  \caption{The density histograms of MER and CS for positive and negative sentence pairs in the training data (Wikipedia and NLI) and the evaluation data (STSB) under various settings.}
  \label{fig:twometrics}
\end{figure*}

\subsection{Match Error Rate and Cosine Similarity}

RFD measures the complexity of the similarity pattern of training data. To intuitively understand the concept of \textbf{pattern complexity}, we select two metrics, Match Error Rate (MER)~\cite{morris_wer_2004} and Cosine Similarity (CS), to reflect the patterns in the data at both the lexical and feature levels. 

Given two sentences $s_1$ and $s_2$, and a sentence encoder $f$ that maps each sentence into an embedding in the hypersphere, the two metrics are defined as follows:
\begin{itemize}
    \item \textbf{Match Error Rate} (MER): this metric quantifies the similarity between two sentences based on edit distance. In calculating this distance, there are four permissible operations: insertion, deletion, substitution, and retain. We use $I, D, S, R$ to represent the count of these four operations, respectively. 
    \begin{align}
        \mathrm{MER}(s_1,s_2) = \frac{I + D + S}{I + D + S + R}. 
    \end{align}
    This metric is calculated at a lexical level, and a lower value means higher similarity;
    \item \textbf{Cosine Similarity} (CS): Cosine similarity, which is assessed in the feature space, is defined as:
    \begin{align}\label{eq:CS}
        \mathrm{CS}(s_1,s_2) = f(s_1)^\top f(s_2),
    \end{align}
    where BERT$_\mathrm{base}$ is utilized as the sentence encoder in our implementation. This metric is calculated at a feature level, and a higher value means higher similarity.
\end{itemize}

We calculate the MER and CS for both positive and negative sentence pairs in the Wikipedia and NLI datasets under all settings. Additionally, we calculate the MER and CS for the STS Benchmark evaluation dataset (STSB for short), where sentence pairs with an STS score higher than 4 are treated as positive sentence pairs, and those with an STS score lower than 1 are treated as negative. We use the density histogram to describe the distribution of these two metrics and plot the results in Figure~\ref{fig:twometrics}. The results are organized into four rows: the first row includes the results for the Wikipedia dataset under three unsupervised settings; the second row includes the results for the NLI training dataset and STSB evaluation dataset, both of which contain the patterns we want to simulate; the third row includes the results for the LLM-generated data whose corpus source is Wikipedia dataset; and the fourth row includes the results for the LLM-generated data whose corpus source is NLI dataset.

To intuitively understand the concept of \textbf{pattern complexity}, we compare the first two rows. It can be found that the distribution of MER in the supervised setting (``NLI'') is broader for both positive and negative sentence pairs compared to that in the unsupervised settings. This implies that the supervised setting contains more complex patterns than the unsupervised settings. Moreover, the distribution of CS in the supervised setting is shifted to the right relative to that in the unsupervised settings for negative sentence pairs, indicating that the negative sentence pairs in the supervised setting are more difficult to identify. The same holds true for positive sentence pairs in most cases. These observations prove that the patterns in the supervised setting are more complex than those in the unsupervised settings, making the patterns in the supervised setting more difficult to fit. It's important to note that RFD is calculated on the same evaluation data. Therefore, when the pattern complexity of the supervised setting is higher than that of the unsupervised setting, the RFD between the supervised setting and the evaluation data is correspondingly higher than the RFD between the unsupervised setting and the evaluation data.

We can also use the MER and CS to intuitively understand the concept of \textbf{pattern simulation}. First, we compare the last two rows with the first row. From the comparison, we can observe that the patterns in the LLM-generated data are more complex than those in the unsupervised settings, confirming that we successfully introduce complex patterns in the generated hybrid datasets. Then, we compare the last two rows with the second row and find that the distribution of data generated by simulating STS patterns (``Wiki.STS'' and ``NLI.STS'') aligns more closely with that of ``STSB'', and the distribution of data generated by simulating NLI patterns (``Wiki.NLI'' and ``NLI.NLI'') aligns more closely with that of ``NLI''. These findings verify that the complex patterns are indeed introduced through pattern simulation.

\subsection{Details of the Prompt}

We adopt the ICL capability of the LLM to simulate the NLI patterns and STS patterns separately. In this subsection, we provide the specific prompts we use:

\noindent\textbf{Simulating NLI patterns} includes two prompts:
\begin{enumerate}
    \item \textit{Generating Positive Sentence} $s_i^\mathrm{p}$:    
    \begin{small}
    \begin{minted}{text}
When the user enters a premise text, plea-
se generate a hypothesis text that stands 
in an entailment relationship to the given 
premise.

In the following illustrative examples, 
the Hypothesis is a logical entailment of 
the Premise:
- Example 1:
  - Premise: Over the years these bags 
    have proved so handy that to keep up 
    with demand the local supply of genu-
    ine handmade bags has been augmented 
    with imported goods.
  - Hypothesis: These genuine handmade 
    bags are very useful and in demand.
- Example 2:
  - Premise: Advocates worry that if subs-
    tantial numbers of welfare mothers are 
    pushed into jobs, centers might be 
    swamped by demands to serve as many as
    a million added kids.
  - Hypothesis: Centers are only capable 
    of serving a limited number of kids.
- Example 3:
  - Premise: A black and white dog is run-
    ning through a snowy field.
  - Hypothesis: A black and white dog is 
    outside.

Generate the hypothesis directly without 
any other interpretation. The generated 
hypothesis should be a logical inference 
from the information available in the pre-
mise. In other words, if the premise is 
true, the hypothesis must also be true.
    \end{minted}
    \end{small}
    \item \textit{Generating Negative Sentence} $s_i^\mathrm{n}$:
    \begin{small}
        \begin{minted}{text}
When the user enters a premise text, 
please generate a hypothesis text that 
presents a contradiction to the informa-
tion provided in the given premise.

In the following illustrative examples, 
The hypothesis logically contradicts the 
Premise:
- Example 1:
  - Premise: Over the years these bags 
    have proved so handy that to keep up 
    with demand the local supply of genu-
    ine handmade bags has been augmented 
    with imported goods.
  - Hypothesis: The bags are known for 
    falling apart and people are looking 
    for an alternative.
- Example 2:
  - Premise: Advocates worry that if sub-
    substantial numbers of welfare mothers 
    are pushed into jobs, centers might 
    be swamped by demands to serve as many 
    as a million added kids.
  - Hypothesis: If welfare mothers get 
    jobs, it decreases the number of kids 
    that centers need to serve.
- Example 3:
  - Premise: A black and white dog is run-
    ning through a snowy field.
  - Hypothesis: The dog is taking a nap.

Generate the hypothesis directly without 
any other interpretation. The hypothesis 
should contradict the information given in 
the premise. This means the premise and 
hypothesis cannot both be true at the same 
time.
        \end{minted}
    \end{small}
\end{enumerate}

\noindent\textbf{Simulating STS patterns} includes three prompts:
\begin{enumerate}
    \item \textit{Generating Positive Sentence} $s_i^\mathrm{p}$:
    \begin{small}
        \begin{minted}{text}
Your task is to generate a new sentence 
that is semantically similar to the user's 
input sentence. 

In the following illustrative examples, 
Sentence 1 and Sentence 2 are semantically 
similar:
- Example 1:
  - Sentence 1: U.S. prosecutors have arr-
    tested more than 130 individuals and 
    have seized more than $17 million in a 
Continuing crackdown on Internet fraud 
    and abuse.
  - Sentence 2: More than 130 people have 
    been arrested and $17 million worth of 
    property seized in an Internet fraud 
    sweep announced Friday by three U.S. 
    government agencies.
- Example 2:
  - Sentence 1: The hearing occurred a day 
    after the Pentagon for the first time 
    singled out an officer, Dallager, for 
    not addressing the scandal.
  - Sentence 2: The hearing came one day 
    after the Pentagon for the first time 
    singled out an officer - Dallager - 
    for failing to address the scandal.
- Example 3: Sentence 2 is generated based 
  on Sentence 1.
  - Sentence 1: The Bush administration 
    blames Hussein loyalists and foreign 
    Muslim militants who have entered Iraq 
    to fight U.S. troops for the wave of 
    bombings and guerrilla attacks.
  - Sentence 2: The Bush administration 
    blames the wave of bombings and guess-
    ill attacks on Saddam loyalists and 
    foreign Muslim militants who have 
    entered Iraq to fight U.S. troops.

Generate the new sentence directly without 
any other interpretation, and make sure it 
maintains the same information as the ori-
original input sentence.
        \end{minted}
    \end{small}
    \item \textit{Generating Intermediate Sentence} $s_i^\mathrm{m}$:
    \begin{small}
        \begin{minted}{text}
Your task is to generate a revised sentence-
ce by omitting certain details in the 
user's input sentence. 

In the following illustrative examples, 
Sentence 2 is created by omitting details 
from Sentence 1:
- Example 1:
  - Sentence 1: Police launched an inter-
National Hunt for Shevaun Pennington 
    after she ran away with a 31-year-old 
    Toby Studabaker Saturday.
  - Sentence 2: Shevaun Pennington 
    disappeared on Saturday morning after 
    arranging to meet 31-year-old Toby 
Studebaker.
- Example 2:
  - Sentence 1: In a news release Thurs-
    day, Strayhorn said this was the 
    first time a comptroller rejected a 
    budget.
  - Sentence 2: Strayhorn said it was the 
    first time in Texas history a computer-
    roller had not certified the appropriate-
actions act.
- Example 3:
  - Sentence 1: The Korean Air deal is 
    expected to be finalized "in the next 
    several weeks," Boeing spokesman Bob 
    Saling said.
  - Sentence 2: Boeing said the final 
    agreement is expected to be signed 
    during the next few weeks.

Generate the revised sentence directly 
without any other interpretation, and 
Make sure that it contains significantly 
fewer details than the original input 
sentence.
        \end{minted}
    \end{small}
    \item \textit{Generating Negative Sentence} $s_i^\mathrm{n}$:
    \begin{small}
        \begin{minted}{text}
Your task is to generate a new sentence 
that conveys a distinct or even contradict-
tory meaning compared to the user's input 
sentence. 

In the following illustrative examples, 
Sentence 2 is generated to convey distinct-
tor contradictory meaning compared to Sen-
tense 1:
- Example 1:
  - Sentence 1: Monkeypox is usually 
    found only in central and western 
    Africa.
  - Sentence 2: Prairie dogs, usually 
    found in southwestern and western 
    states aren't indigenous to Wiscon-
    sin.
- Example 2:
  - Sentence 1: The tech-laced Nasdaq Com-
    posite Index gained 2.90 points, or 0.
    18 percent, to 1,606.87.
  - Sentence 2: At 12:10 p.m. EDT, Cana-
    da's benchmark S&P/TSX composite index 
    was up 6.87 points or 0.1 percent to 
    6,979.29.
- Example 3:
  - Sentence 1: A man is playing the dr-
    ums.
  - Sentence 2: A woman is slicing some 
    leaves.

Generate the new sentence directly without 
any other interpretation.
        \end{minted}
    \end{small}
\end{enumerate}

Every prompt has three examples randomly sampled from the pattern source, and the examples are fixed for every pattern simulation.



\end{document}